# Fast and Accurate 3D Medical Image Segmentation with Data-swapping Method

Haruki Imai[1], Samuel Matzek[2], Tung D. Le[1], Yasushi Negishi[1], and Kiyokuni Kawachiya[1]

[1] IBM Research, Tokyo, Japan
{imaihal, tung, negishi, kawatiya}@jp.ibm.com
[2] IBM Systems, MN, USA
smatzek@us.ibm.com

**Abstract.** Deep neural network models used for medical image segmentation are large because they are trained with high-resolution three-dimensional (3D) images. Graphics processing units (GPUs) are widely used to accelerate the trainings. However, the memory on a GPU is not large enough to train the models. A popular approach to tackling this problem is patch-based method, which divides a large image into small patches and trains the models with these small patches. However, this method would degrade the segmentation quality if a target object spans multiple patches. In this paper, we propose a novel approach for 3D medical image segmentation that utilizes the data-swapping, which swaps out intermediate data from GPU memory to CPU memory to enlarge the effective GPU memory size, for training high-resolution 3D medical images without patching. We carefully tuned parameters in the data-swapping method to obtain the best training performance for 3D U-Net, a widely used deep neural network model for medical image segmentation. We applied our tuning to train 3D U-Net with full-size images of 192 × 192 × 192 voxels in brain tumor dataset. As a result, communication overhead, which is the most important issue, was reduced by 17.1%. Compared with the patch-based method for patches of 128 × 128 × 128 voxels, our training for full-size images achieved improvement on the mean Dice score by 4.48% and 5.32 % for detecting whole tumor sub-region and tumor core sub-region, respectively. The total training time was reduced from 164 hours to 47 hours, resulting in 3.53 times of acceleration.

**Keywords:** Deep Learning, Image Segmentation, 3D U-Net, Data-swapping Method.

## 1 Introduction

Medical image analysis of magnetic resonance imaging (MRI) and computed tomography is important for early detection of diseases, appropriate treatment planning, surgical planning, and prognostic observation. Medical image segmentation is a key analysis for detecting lesion boundaries. The segmentation is processed manually by pathologists, but manual segmentation is subjective and time-consuming because it may be affected by the pathologists' bias, and it is necessary to carefully analyze multimodal three-dimensional (3D) images simultaneously [1] [2]. The segmentation quality also depends on the pathologists' experience. Therefore, automatic segmentation is highly desired.

Deep learning is widely used to automate and aid medical image segmentation. The number of scientific papers on deep learning in medical image segmentation rapidly increased in 2015 and 2016, and the topic is now dominant [3]. Deep neural network models for segmentation are also large because they are trained with multimodal high-resolution 3D images.

Most neural network models for medical image segmentation are trained on graphics processing units (GPUs) to accelerate performance. However, GPU memory is not large enough to train these neural network models for segmentation. To meet such demands, GPU memory capacity has been increased to 32 GB in NVIDIA's latest GPU, Tesla® V100, but it is not expected to increase drastically in the future because expensive memory, called "high bandwidth memory," is used for GPUs to achieve a high memory-access throughput. Therefore, GPU memory capacity has become a serious problem for processing large neural network models.

To reduce memory consumption, methods of partitioning an original image into small patches are frequently used [4] [5] [6]. However, the small patches are impossible to capture large lesions that span multiple patches, leading to degradation in segmentation quality. Therefore, it is preferable to use the original full-size image as it is.

Given the fact that central processing unit (CPU) memory is often larger than GPU memory, one of the method for avoiding image partitioning is the data-swapping method, which eases the burden of GPU memory by swapping out intermediate data from GPU to CPU memory while they are not necessary for current GPU computation. Another method is called re-computation. This method discards intermediate data once instead of swapping data, and then, computes them again from some checkpoints when necessary. Compared with the data-swapping method, it requires additional computation. Therefore, the training time becomes longer. There are some studies on the data-swapping method [7] [8] and re-computation method [9], but, there are no studies on medical image segmentation.

In this paper, we propose a novel approach for 3D medical image segmentation that utilizes the data-swapping method to avoid image partitioning into multiple small patches. Our contributions in this paper are as follows:

- We show, for the first time, that medial image training with 3D U-Net [10] becomes possible by using the data-swapping method. By tuning parameters in the data-swapping method, communication overhead, which is most important issue, was reduced by 17.1 %. The data-swapping method with parameter-tuning was 14.4% faster than the re-computation method in training time for one epoch.
- We confirm that our approach improves the mean Dice scores of tumors by 4.48% in the whole tumor sub-region and by 5.32% in the tumor core sub-region compared with the patch-based method.
- We confirm that our approach also accelerates the total training time by about 3.53 times compared with the patch-based method.

The rest of this paper is organized as follows. In Section 2, we describe related work. In Section 3, we give details on the data-swapping method and parameter tuning for 3D U-Net. In Section 4.1, we describe a model and dataset for our experiments. In Section 4.2 and 4.3, we evaluate training time and segmentation quality. In Section 5, we discuss performance comparison with the re-computation method. Finally, we conclude this study.

## 2    Related Work

There are studies on the patch-based method that introduce additional techniques for obtaining the global features of an image because a small patch cannot capture them. Nazeri et al. [6] used a patch-wise convolutional neural network (CNN) and image-wise CNN. The patch-wise CNN first trains local features; then, the image-wise CNN trains the global features by using the feature maps generated from the patch-wise CNN. Hou et al. [5] first used a CNN to detect discriminative patches and then trained an image-level decision fusion model by using histograms of patch-level prediction. However, studies on 3D U-Net [10] [11] have not used such additional techniques. Therefore, we also did not apply them in this paper for comparison.

There are two approaches to running large neural network models on GPUs. The first approach involves using the data-swapping method which is proposed in this paper. M. N. Rhu et al. [7] and Meng et al. [8] also used this approach. They used popular neural networks such as ResNet50 for evaluation and basically focused on the increase in batch size. However, an increase to a very large batch size is not always practical because it does not always provide better results [12]. In this paper, we apply an implementation in TensorFlow for medical image segmentation and show better segmentation quality. To the best of our knowledge, this is the first medical imaging example with the data-swapping method.

The second approach involves using the re-computation method. This is a method of re-computing intermediate data when necessary instead of swapping data. In the forward propagation process, the intermediate data are discarded once. In backpropagation, the forward propagation process is executed again to calculate them. With this method, the overhead of re-computation becomes large, especially when the model is deep, so there is a method of reducing the overhead by holding some intermediate data, which is called a "checkpoint" [9]. Re-computation can be done from checkpoints, so the amount of calculation can be reduced. Compared with the data-swapping method, this does not require CPU-GPU communication, but it consumes more GPU memory and requires additional computation instead. Therefore, the maximum model size becomes smaller than the data-swapping method, and additional overhead occurs due to computation. We discuss them in Section 5.

## 3 Data-swapping Method

### 3.1 Training with Data-swapping Method

In this section, we describe the data-swapping method. **Fig. 1.** shows the processing and data flow of the training phase of an L layer neural network. An intermediate result $A^l$, which is generated in layer $l$ in the forward propagation process, is reused by the same layer in the backward propagation process. The $A^i$(s), $1 \leq i \leq L$, are called feature maps, and they consume large amounts of GPU memory because they are kept in GPU memory until the backward propagation process is finished. However, the data-swapping method swaps out the feature maps from GPU memory to CPU memory after the forward propagation process is finished and then swaps them from CPU memory to GPU memory before executing the layer in the backward propagation process. Because the feature maps are usually large, swapping them out can greatly reduce GPU memory.

With the data-swapping method, CPU memory can be effectively used. However, this method introduce an overhead in transferring data between GPU memory and CPU memory. Reducing these communication overheads is a challenge for the data-swapping method. We reduced these overheads by tuning the parameters described in Section 3.3.

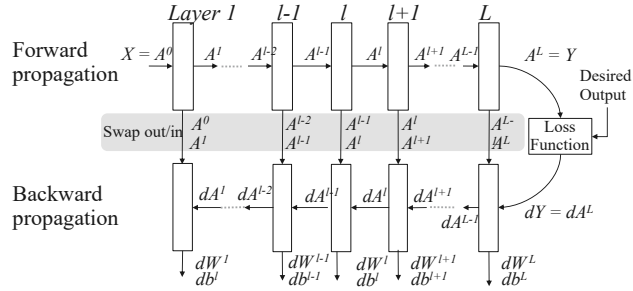

Fig. 1. Processing and data flow of L layers

### 3.2 Implementation in TensorFlow

There is an implementation of the data-swapping method for TensorFlow, called "TensorFlow Large Model Support" (TFLMS) [13], which is published in a GitHub repository [14]. In TensorFlow, neural network models defined by users are internally transformed into a computational graph. TFLMS searches the graph to find edges to insert swap-out/in nodes, and then automatically modifies the graph by inserting nodes for swapping-out and swapping-in.

Swapping all of the feature maps out/in introduces a large communication overhead. Therefore, TFLMS provides parameters for minimize the communication overhead of data swapping. The first important parameter, called "*n_tensors*," allows us to specify the number of feature maps to be swapped out. It searches the graph using breadth-first search and counts the number of feature maps to be swapped out. This parameter is effective because the feature maps of earlier layers are usually large and are kept in GPU memory for a long time. Hence, they should be swapped out first. The second important parameter, called "*lb*," controls how soon the data are swapped back in before use. Since feature maps are swapped-in earlier by increasing *lb*, the communication of swapping-in and GPU computation can overlap. Besides, there are parameters for selecting specific feature maps to be swapped out manually, called "*excl_scopes*" and "*incl_scopes*." Specific feature maps can be selectively swapped by listing the names of the feature maps. To reduce communication overhead, we should choose the best combination of these parameters, considering the characteristics of neural network models.

### 3.3 Parameter Tuning Strategy for 3D U-Net

**Fig. 2** shows four configurations of parameter tuning for 3D U-Net. For each configuration, the left side shows 3D U-Net's architecture and the right side illustrates the results of timelines of one training iteration including one forward propagation and one backward propagation. The architecture has an analysis and a synthesis path, and they are connected with shortcut connections. Each box represents a feature map. The feature maps that have outlined boxes are swapped in each configuration. In the timelines, each box shows GPU computation time to generate a feature map, or communication time for swapping in/out a feature map.

In configuration 1, we swapped all of the feature maps. Although this reduces memory consumption largely, but the communication overhead becomes large. The blanks in GPU computation show the communication overhead. There is large overhead occurred between the end of forward propagation and the start of backward propagation. In the forward propagation, a swapping-out issues after a layer producing a feature map. Since there are many swapping-outs, their communications are not completed during forward propagation, the backward propagation needs to wait until they are completed. The other overheads appear in the backward propagation because layers need to wait for the communication for swapping-ins. In configuration 2, we swapped only the first N feature maps by using the TFLMS parameter *n_tensors* in order to reduce the number of swappings. In this configuration, TFLMS automatically decide swapped feature maps. However, in 3D U-Net, latter feature maps in the synthesis paths are selected because breadth-first search is used to count the number of feature maps. Therefore, this configuration is not enough to remove the overhead between forward propagation and backward propagation. In configurations 3 and 4, we manually avoid swapping feature maps on the synthesis path by using the parameter *excl_scopes*. This allows removing the overhead by swapping out only feature maps in the earlier phase of the forward propagation. Configuration 4 further removes the other overheads during the backward propagation by making swapping-ins happen earlier by using parameter, *lb*.

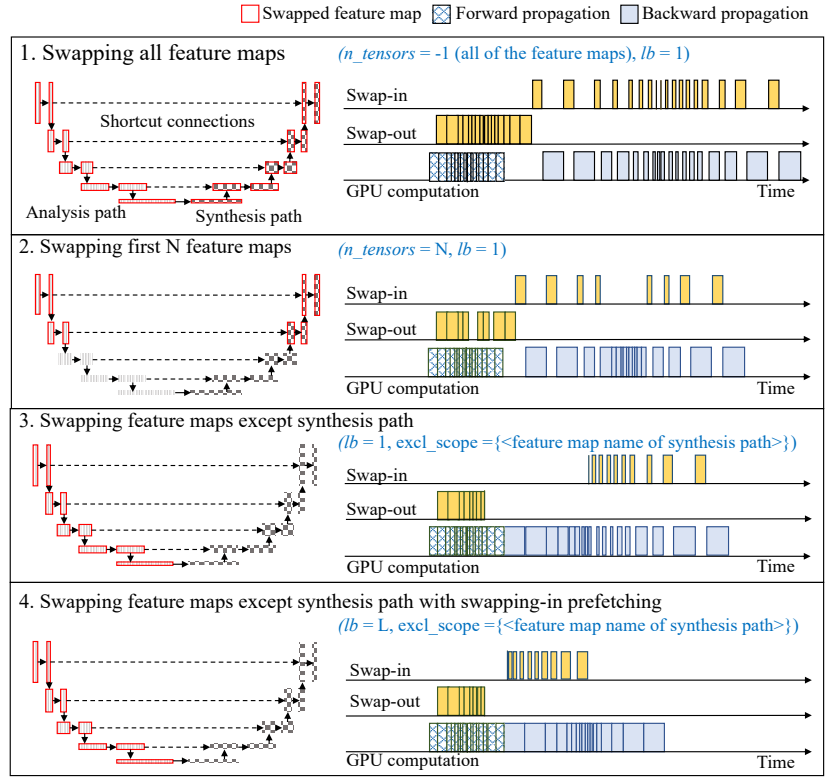

**Fig. 2.** Parameter tuning for 3D U-Net: swapped feature maps in 3D U-Net (left) and timeline of GPU computation and CPU-GPU communication (right)

## 4 Experiments and Results

### 4.1 Model and Dataset

We used a Keras model for 3D U-Net written by David G. Ellis [15]. This model was implemented to process multimodal MRI data following the model architecture described by Isensee et al. [16], which received 3rd place in the Multimodal Brain Tumor Image Segmentation Benchmark (BRATS) 2017 challenge [17]. The model was inspired by the U-Net architecture, and new modules were added such as instance normalization [18] instead of batch normalization. The model used Dice loss as an objective function, which is based on the Dice coefficient [19]. It was written in expectation of a Theano backend for Keras. We modified the code to work with the Keras APIs included in TensorFlow 1.8. The TFLMS Keras callback was then added to the list of callbacks.

The dataset for training, validation, and testing was the official 2017 BRATS challenge dataset [20], which includes segmentation labels. There are 285 images of 285 subjects in total, in which 171 images are used for training, 57 images are used for validation, and the other 57 images are used for testing. The maximum image size in the dataset is 159 × 191 × 151. When training the model by using the full images, memory usage exceeds the GPU memory capacity. Therefore, the patch-based method was used in the studies of David G. Ellis [15] and Isensee et al. [16] by partitioning the images into patches of 128 × 128 × 128 voxels. We used the data-swapping method to use the full images and compared the results with the patch-based method.

The training was done by using the ADAM optimizer with an initial learning rate of $5 \cdot 10^{-4}$, learning rate drop factor of 0.5, and patience of 10, which means that the learning rate dropped by this factor when the validation loss did not improve for 10 epochs. We stopped the training when the validation loss did not improve for 50 epochs by using an early stopping function to prevent overfitting. We used 5-fold cross-validation to validate the results.

Since the number of MRI data is usually limited, data-augmentation techniques are important in medical image segmentation to prevent overfitting [16]. We applied random flip of the axis and permuting in various directions before every training iteration. Since the augmentations run on a CPU, they ran in parallel with the training iterations. Therefore, there is usually no issue with additional overhead. However, if one training iteration on the GPU becomes small, it can cause an overhead in training time.

For the evaluations of training time and segmentation quality discussed in the following sections, we used IBM® Power Systems™ S822LC for High Performance Computing, which has two POWER8 CPUs (10 cores with 3.54 GHz) and 512 GB of CPU memory. There are four NVIDIA® Tesla® P100s, each of which had 16-GB of GPU memory. We used only one GPU for all evaluations. The CPU and GPU were connected with NVLink 1.0, which has a bidirectional bandwidth of 80 GB/sec. The OS was RHEL 7.3, and we used CUDA9.1, and cuDNN7.0.2. As for the data-swapping method, we used TensorFlow 1.8 with TFLMS [14](commit #fe05c31).

### 4.2 Acceleration of Training Time

**Results of Parameter Tuning.**

In this section, we describe how much parameter tuning of the data-swapping method accelerated training time. **Fig. 3** and **Table 1** show the results of the parameter tuning. **Fig. 3** shows the training time of one epoch. Compared with configurations 1 and 4, the training time was improved by 17.1%. As shown in **Table 1**, the total number of feature maps was 843. In configurations 3 and 4, since several feature maps in the synthesis path were manually set in *excl_scopes,* the number of swapped feature maps was smaller than that of configuration 1. In terms of peak memory usage, that of configuration 1 was small because all of the feature maps were swapped, but in other configurations, it increased when the number of swapped feature maps was reduced.

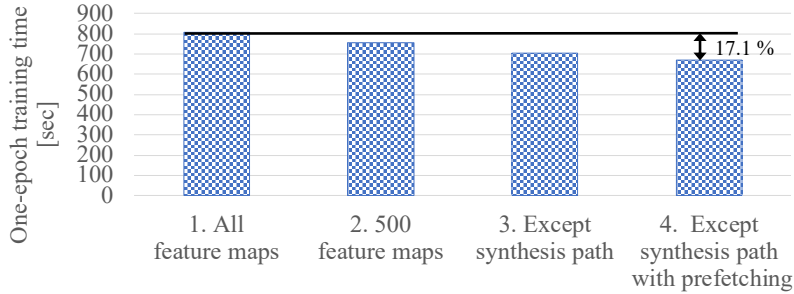

**Fig. 3.** Results of parameter tuning

**Table 1.** TFLMS configuration and results: feature maps of synthesis path are specified in *excl_scopes* in configurations 3 and 4. In *n_tensors* = −1, all of the feature maps are swapped after excluding feature maps specified in *excl_scopes*.

| Configuration | | 1 | 2 | 3 | 4 |
|---|---|---|---|---|---|
| TFLMS Configu-ration | *n_tensors* | −1 | 500 | −1 | −1 |
| | *lb* | 1 | 1 | 1 | 20 |
| | *excl_scopes* | | | ✓ | ✓ |
| #swapped feature maps | | 843 | 500 | 599 | 599 |
| Peak memory usage [GB] | | 12.5 | 13.5 | 14.2 | 13.5 |

**Comparison with Patch-based Method.**
In this section, we compared the training time of our approach with the patch-based method. **Fig. 4** shows one-epoch training time and total training time. The total training times were training time for which the results of 5-fold cross-validation were averaged. We used the maximum batch size for each method, which was 2 for the patch-based method and 1 for our approach. The training times were greatly accelerated by avoiding image partitioning. They were 2.98 times faster in one epoch. The total training time is the time until training automatically stopped through the early stopping function. The total training time on our approach was 3.53 times faster than that on the patch-based method. The ratios were different from one-epoch training time because the total number of epochs between the patch-based method and our approach was different.

In the patch-based method, a 3D image is partitioned into multiple patches, which means multiple training iterations are required to train one image. This introduces additional computation. Our approach requires 171 iterations because there are 171 images for one epoch and they are not partitioned. The number of training iterations in the patch-based method was 387. Also, since computation for one iteration becomes smaller in small patch size, pre-processing such as data-augmentation running on a CPU becomes relatively larger, so CPU computation becomes a bottleneck.

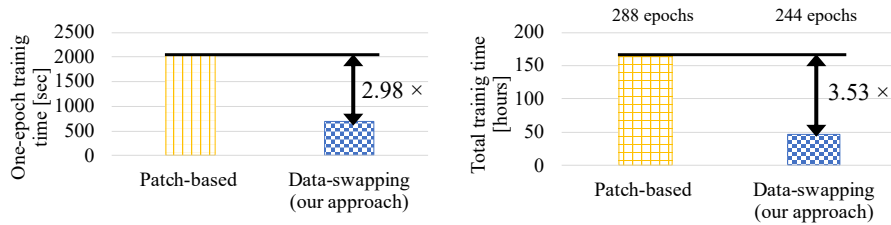

**Fig. 4.** One-epoch training time (left) and average total training time (right)

### 4.3 Improvement of Segmentation Quality

**Quantitative Evaluations.**

After training was finished, we evaluated the segmentation quality by using different datasets for testing. Four structures, i.e., edema, non-enhancing solid core, necrotic/cystic core, and enhancing core, were labeled in the dataset. During the evaluations, we grouped the structures into three sub-regions to represent better practical clinical applications, that is, whole tumor, tumor core, and enhancing tumor, which are the same metrics as those in the BRATS challenge. The whole tumor sub-region includes all four tumor structures, the tumor core sub-region includes all tumor structures except the edema, and the enhancing tumor sub-region includes only the enhancing core.

For quantitative evaluation of the tumor sub-regions, we calculated the Dice score, sensitivity (true positive rate), and specificity (true negative rate), which are widely used in image segmentation [21] and the BRATS challenge. They are calculated as follows.

$$\text{Dice}(P, T) = \frac{|P_1 \wedge T_1|}{(|P_1| + |T_1|)/2}$$

$$\text{Sensitivity}(P, T) = \frac{|P_1 \wedge T_1|}{|T_1|}$$

$$\text{Specificity}(P, T) = \frac{|P_0 \wedge T_0|}{|P_0|}$$

, where P represents the prediction results, and T represents the ground truth labels. $T_1$ and $P_1$ are the subsets of voxels predicted as positives for the tumor region, and $T_0$ and $P_0$ are those predicted as negative. We have $P = P_0 + P_1$ and $T = T_0 + T_1$. **Fig. 5** shows box plots of the Dice score of the whole tumor, tumor core, and enhancing tumor. This is one of the training results of 5-fold cross-validation. **Table 2** lists the mean scores of each metrics of 57 subjects. They are averages of 5-fold cross-validation. Compared with the results of the patch-based method, the Dice score of our approach improved by 4.48 % for the whole tumor sub-region and by 5.32 % for the tumor core sub-region. This indicates that our approach achieved better scores compared with the patch-based method for the whole tumor and tumor core sub-regions. However, the result of enhancing tumor was similar to the patch-based method. This is because an enhancing tumor is basically smaller than the whole tumor and tumor core, and a small patch sizes are large enough to capture them.

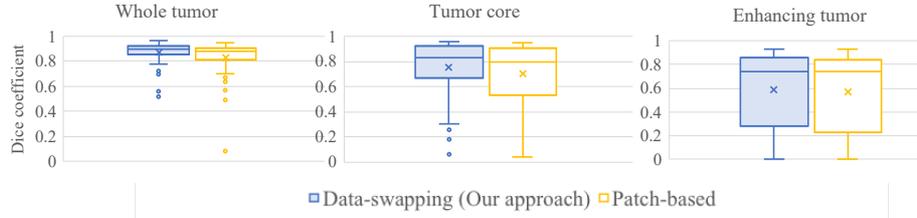

**Fig. 5.** Box plots of Dice score of each tumor sub-region

**Table 2.** Mean scores of each metrics (average of 5 training results)

| Method | Dice | | | Sensitivity | | | Specificity | | |
|---|---|---|---|---|---|---|---|---|---|
| | Whole | Core | En-hancing | Whole | Core | En-hancing | Whole | Core | En-hancing |
| **Data-swapping (Our approach)** | **0.865** | **0.771** | **0.641** | **0.870** | **0.799** | **0.675** | **0.966** | **0.969** | **0.971** |
| Patch-based | 0.826 | 0.730 | 0.637 | 0.826 | 0.758 | 0.671 | 0.963 | 0.969 | 0.971 |

**Visualized Evaluations.**
We visualized the results in **Fig. 6**. We selected a slice from four subjects. The figures of each subject show the original image, ground truth (GT), the results of our approach, and that of the patch-based method. We captured a slice of a 3D image that clearly expresses the difference. The slices shown are the results from the axial view. Yellow denotes "edema," green denotes "necrotic/cystic core," and blue denotes "enhancing core." The results of our approach were more similar to the ground truth than that of the patch-based method as a whole. Wrong detections tended to be observed outside the tumor. This is because a small patch cannot cover all of a tumor at once.

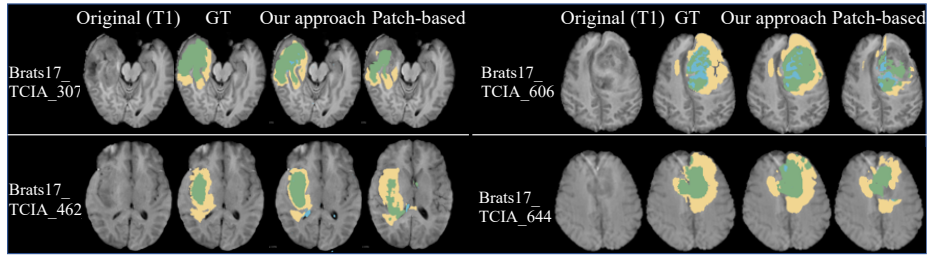

**Fig. 6.** Visual results of segmentation (one of the results of 5-fold cross-validation)

4## 5 Discussion

In this section, we discuss performance comparison with re-computation method, which is another approach to avoiding image partitioning into small patches described in Section 2. There is an implementation of the re-computation method for TensorFlow called "gradient-checkpointing" [22]. This enables the re-computation by replacing gradients function. For performance tuning, it provides a parameter to choose which types of operation as checkpoints. We used a parameter called *speed*, which enables small overhead by choosing feature maps of convolutions and matrix multiply operations as the checkpoints. Since the operations are usually computation-intensive operations, the amount of additional computation is reduced.

Since our approach and re-computation method train the same model with different execution methodology, the segmentation quality is similar and the difference appears only in training time. Table 3 shows that our approach was 14.4 % faster than the re-computation method in one-epoch training time. In our experiments, we used a machine with a fast CPU-GPU connection. This should contribute to reducing the overhead produced with our approach. When using a slow CPU-GPU connection such as PCI-e (32 GB/sec), hybrid use of data-swapping and re-computation may be suitable [23].

The re-computation method keeps the checkpoints on GPU memory, which consumes larger memory consumption than the data-swapping method. Actually, the re-computation method could not train dataset of 208 × 208 × 208 voxels, but our approach was able to do it. This is another advantage of our approach.

Table 3. One-epoch training time [sec]

| | |
|---|---|
| Re-computation | 783 |
| Data-swapping (Our approach) | 670 |

## 6 Conclusion

We proposed a novel approach for 3D medical image segmentation that utilizes the data-swapping method to avoid image partitioning into multiple small patches. We showed that our approach made it possible to train 3D U-Net using a brain tumor dataset without partitioning into small patches and improved the mean Dice scores of tumors by 4.48% in the whole tumor sub-region and by 5.32% in the tumor core sub-region. Also, our approach accelerated the training time by 3.53 times. The most important issue in our approach is the communication overhead, but we carefully tuned parameters to minimize the overhead in consideration of the characteristics of the 3D U-Net and the overhead was reduced by 17.1 %. In the future, we will apply our approach to other medical image segmentation with larger images. That may incur larger communication overhead, which the parameter tuning cannot reduce fully. In that case, hybrid use with re-computation method should be effective.